\documentclass[runningheads]{llncs}

\usepackage[mobile]{eccv}
\usepackage{eccvabbrv}

\usepackage{graphicx}

\usepackage{tikz}
\usepackage{comment} 
\usepackage{amsmath,amssymb} 
\usepackage{color}
\usepackage{bm}
\usepackage{caption}

\usepackage[pagebackref,breaklinks,colorlinks,citecolor=eccvblue]{hyperref}

\usepackage{graphicx}
\usepackage{booktabs}

\usepackage[accsupp]{axessibility}  

\usepackage{hyperref}

\begin{document}

\title{TELA: Text to Layer-wise 3D Clothed Human Generation} 

\titlerunning{TELA: Text to Layer-wise 3D Clothed Human Generation}

\author{Junting Dong\inst{1} \and
Qi Fang\inst{2} \and
Zehuan Huang\inst{3} \and
Xudong Xu\inst{1} \and \\
Jingbo Wang\inst{1} \and
Sida Peng\inst{4} \and
Bo Dai\inst{1}}
\authorrunning{J. Dong et al.}
\institute{$^1$ Shanghai AI Laboratory 
$^2$ NetEase Games AI Lab \\ $^3$ Beihang University $^4$  Zhejiang University \\
\url{http://jtdong.com/tela_layer/}}

\maketitle

\begin{abstract}
This paper addresses the task of 3D clothed human generation from textural descriptions. 
Previous works usually encode the human body and clothes as a holistic model and generate the whole model in a single-stage optimization, which makes them struggle for clothing editing and meanwhile lose fine-grained control over the whole generation process. 
To solve this, we propose a layer-wise clothed human representation combined with a progressive optimization strategy, which produces clothing-disentangled 3D human models while providing control capacity for the generation process. 
The basic idea is progressively generating a minimal-clothed human body and layer-wise clothes. 
During clothing generation, a novel stratified compositional rendering method is proposed to fuse multi-layer human models, and a new loss function is utilized to help decouple the clothing model from the human body. 
The proposed method achieves high-quality disentanglement, which thereby provides an effective way for 3D garment generation. 
Extensive experiments demonstrate that our approach achieves state-of-the-art 3D clothed human generation while also supporting cloth editing applications such as virtual try-on.


\end{abstract}    

\section{Introduction}
\label{sec:intro}


The generation of 3D clothed human is of great need in a variety of applications such as AR/VR, immersive telepresence, and virtual try-on.
In these applications,
the generation process is hoped to be highly controllable and detachable,
where the human model and its set of clothes can be created independently,
and simultaneously support free clothing replacement and transfer.
This requires a faithful disentanglement of the human model and clothes. 
While manually generating such decomposed clothed human can be labor-intensive and time-consuming,
auto-generation conditioned on inputs from complex capture systems (e.g., 3D scans \cite{saito2021scanimate} and multi-view videos \cite{collet2015high,guo2019relightables} is also unscalable and inaccessible.
In real applications users often desire the clothed human can be created with simple inputs, such as textual descriptions.

\begin{figure*}[t]
\centering
\includegraphics[width=1\linewidth,trim={0cm 82cm 0cm 0cm},clip]{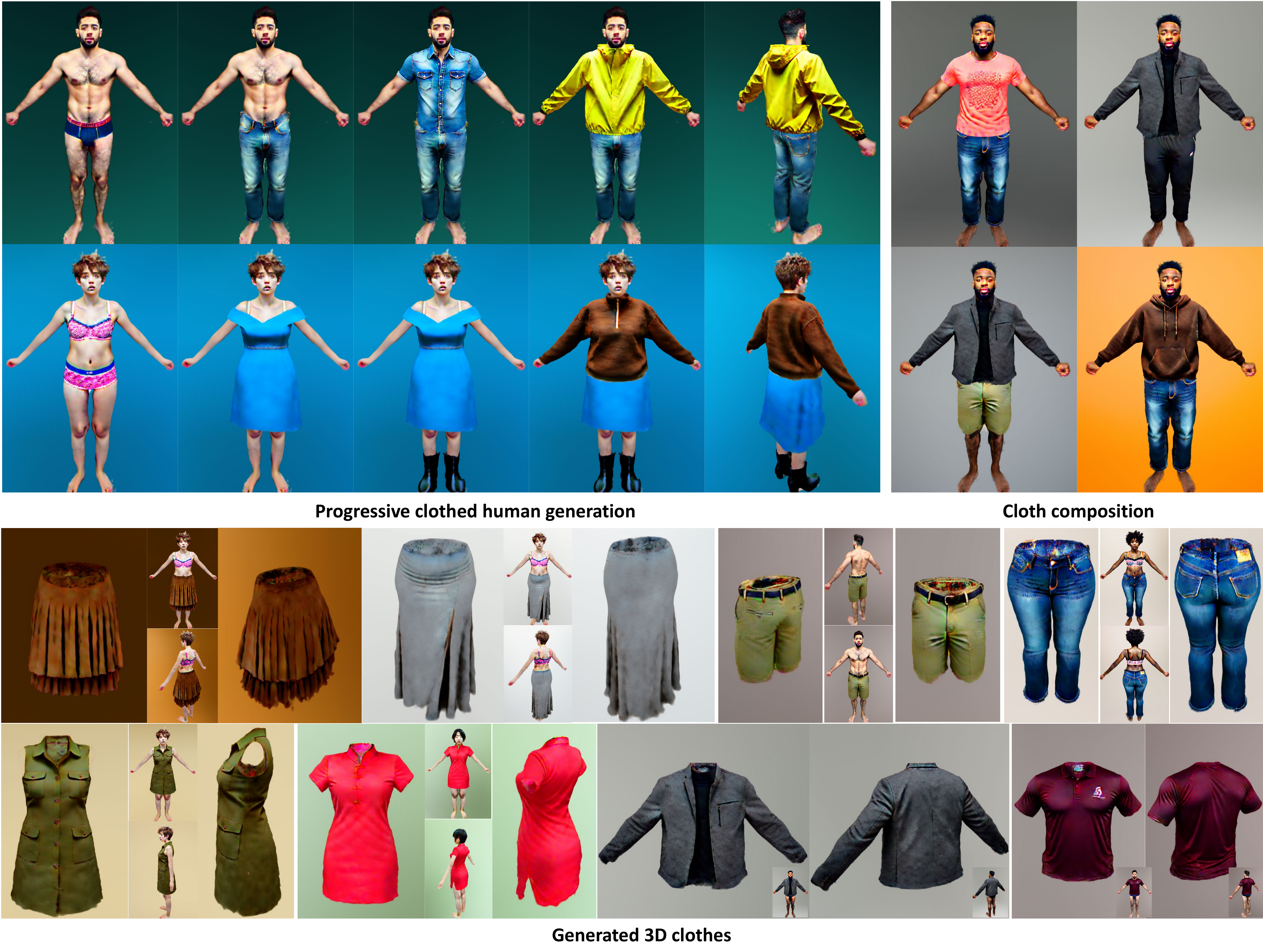} 
\captionof{figure}{Given textural descriptions (e.g., ``a man wearing jeans, a denim shirt, and a windbreaker''), this paper aims to generate clothing-disentangled 3D human models progressively. Meanwhile, our approach enables high-quality 3D cloth generation and supports applications like cloth composition.  }
\label{fig:teaser}
\end{figure*}

Benefiting from the rapid development of Large Language Models \cite{2020t5,brown2020language} and Diffusion Models \cite{rombach2022high,ramesh2022hierarchical,sr3}, some works have begun to explore various text-to-3D tasks. Recently, DreamFusion \cite{poole2022dreamfusion} proposes to leverage a pretrained text-to-image diffusion model to generate 3D objects in an optimization-based pipeline, under the guidance of Score Distillation Sampling (SDS).
However, due to the lack of human prior, they still struggle for high-quality clothed human generation. To address this, some works \cite{cao2023dreamavatar,kolotouros2023dreamhuman,huang2023dreamwaltz} propose to combine the linear blending skinning algorithm \cite{smpl} into the optimization-based pipeline, which significantly improves the quality of generated clothed humans. However, these methods usually represent the human model and its set of clothes as a holistic model and generate the whole model in a single-stage optimization.
This not only makes clothing editing such as clothing replacement infeasible,
but also results in the loss of control over the whole generation process (e.g., can not specify the order of inside and outside of clothes ) and thereby limits their applicability.



%

In this paper, we propose TELA,  a novel approach for the new task of clothing disentangled 3D human model generation from texts. 
The proposed approach introduces a layer-wise clothed human representation, where the human body and each clothing are represented with separate neural radiance fields (NeRFs) \cite{mildenhall2020nerf}. 
Then, a progressive optimization strategy is proposed to generate the minimal-clothed human body and layer-wise clothes sequentially, which effectively outputs a disentangled clothed human model and also provides the users sufficient control of the generation process, as shown in Figure~\ref{fig:teaser}.

However, achieving high-quality decoupled clothing generation is challenging. 
First, existing multi-layer representations \cite{ranade2022ssdnerf,wu2023dorec} commonly define all NeRF layers in the same space and then adopt the \textit{Max} or \textit{Mean} operation to composite multi-layer outputs, which may lead to penetration between adjacent layers as shown in the experiments. 
It is more reasonable to define each NeRF in its own layer space. However, how to divide areas of different layers is unknown. 
Second, unlike those disentangled reconstruction methods that take accurate part masks as inputs, we aim to achieve the disentanglement with only texts, which makes the task more difficult. 
Our experiments show that only using body-level SDS loss\cite{cao2023dreamavatar,kolotouros2023dreamhuman,huang2023dreamwaltz} cannot obtain high-quality cloth decoupling.

To address these challenges, we first propose a novel stratified composition method. 
During volume rendering, this method can automatically divide areas of different layers according to the transparency of each point along each ray. 
On the divided area of a specific layer, only the corresponding NeRF is utilized to produce the density and color, which avoids the interpenetration between adjacent layers. 
For cloth disentanglement, we propose novel dual SDS losses to simultaneously supervise the rendered clothed human image and clothes-only image, which introduces more regularization on the cloth model and encourages it to decouple from the human body. 

Extensive experiments demonstrate that TELA effectively disentangles the human body and each cloth. Thanks to the disentangled modeling, our approach supports clothing editing applications such as virtual try-on, which is hard to achieve in prior methods. It is worth mentioning that, from another perspective, our paper presents a novel method for high-quality 3D clothing generation by simultaneously considering the underlying human body.

In summary, this work makes the following contributions:
\begin{itemize}

\item We present a novel framework for the new task of cloth disentangled 3D human generation from textual inputs, which also provides an effective way for 3D garment generation.


\item We propose a novel decoupled cloth generation algorithm that introduces stratified composition method for multi-layer rendering and dual SDS losses for cloth disentanglement.


\item Compared to the holistic modeling method, our approach achieves better clothed human generation quality while the disentangled modeling unleashes the potential of many downstream applications (e.g., virtual try-on).


\end{itemize}

\section{Related works}
\label{sec:formatting}

\paragraph{Text-guided 3D content generation.}

Capitalizing on the significant advancements in Text-to-Image (T2I) generation models \cite{dalle2,rombach2022high,imagen}, numerous studies have delved into the domain of text-to-3D generation. Certain approaches \cite{nichol2022point,jun2023shap} advocate for the development of a text-conditioned 3D generative model using paired 3D and text data. Nevertheless, the limited size of existing paired 3D-text datasets, compared to their image-text counterparts \cite{laion5b}, constrains the generalization capability of these methods. Consequently, many works have investigated the use of pre-trained 2D generation models to generate 3D content without any 3D data. Early works propose to leverage the CLIP model \cite{clip} to optimize underlying 3D representations (e.g., neural radiance fields \cite{jain2021dreamfields} or meshes \cite{clipmesh}) by minimizing the embedding difference between rendered images and accompanying text descriptions. Owing to the limited capacity of the CLIP space, these endeavors often result in the generation of unrealistic 3D models. Recently, \cite{poole2022dreamfusion,lin2022magic3d} propose the utilization of powerful text-to-image diffusion models for optimization with Score Distillation Sampling (SDS), leading to impressive results. Furthermore, \cite{wang2023prolificdreamer} suggests a particle-based variational approach, namely variational score distillation, to enhance the quality of 3D generation. Despite their success in general object generation, these methods still face challenges in achieving a high-quality generation of clothed human models.


\paragraph{Text-guided 3D human generation.} 
Building upon the foundation of 2D pre-trained generative models, several studies aim to enhance 3D human generation by incorporating a human prior. Avatar-CLIP \cite{avatarclip} enhances the parametric human model SMPL \cite{smpl} by integrating the neural radiance field as the 3D human representation, followed by optimization using the CLIP model. Likewise, leveraging the parametric human model as a human prior, some works \cite{cao2023dreamavatar} utilize pre-trained text-to-image diffusion models and Score Distillation Sampling for 3D human generation. DreamHuman \cite{kolotouros2023dreamhuman} introduces a deformable and pose-conditioned NeRF model, achieving animatable 3D clothed human generation. Recently, to alleviate the Janus (multi-face) problem, DreamWaltz \cite{huang2023dreamwaltz} introduces 3D-consistent score distillation sampling. The approach involves projecting the canonical 3D human skeleton to each view and adopting a 2D human skeleton-conditioned diffusion model \cite{controlnet} for view-aligned optimization, yielding impressive results. AvatarVerse \cite{zhang2024avatarverse} replaces the human skeleton condition with the densepose map. Moreover, TADA \cite{liao2023tada} replaces the NeRF model with a deformable SMPL-X \cite{smplx} mesh and a texture map, seamlessly integrating the results into existing Computer Graphics workflows. HumanNorm \cite{huang2023humannorm} proposes to fine-tune the diffusion model to generate normal maps and achieve remarkable human geometry generation. GAvatar \cite{yuan2023gavatar} introduces the Gaussian splatting representation for efficient and high-quality human rendering. While these methods demonstrate remarkable success in clothed human generation, they commonly represent the clothed human as a holistic model, neglecting the layer-wise nature and facing challenges in cloth editing.


\paragraph{Clothed human modeling.} 

Early methods \cite{alldieck2018video,bhatnagar2019multi,ma2020learning} commonly depend on parametric human meshes and incorporate additional vertex offsets for clothed human modeling. Benefiting from advancements in implicit functions \cite{mescheder2019occupancy,park2019deepsdf,mildenhall2020nerf}, recent methods \cite{saito2019pifu,peng2021neural,peng2021animatable,dong2022totalselfscan,saito2021scanimate,tiwari2021neural,MetaAvatar,xiu2022icon,chen2022gdna} have presented impressive clothed human reconstruction or generation from images and 3D scans. Typically, these approaches treat the human body and clothing as an integrated entity. To disentangle the representation of the human body and clothing, some studies \cite{Yu2018DoubleFusionRC,PonsMoll2017ClothCapS4,Yu2019SimulCapS,chen2021tightcap,Feng2022CapturingAA} propose a multi-layer human representation, achieving high-quality reconstruction from 3D scans, depth sensors, and even monocular videos. These techniques rely on a parametric human model \cite{smpl} and then reconstruct separate layers for clothing. In addition, there are some works proposing the learning of 3D generative clothed models from 3D garment datasets. BCNet \cite{jiang2020bcnet} proposes the learning of parametric mesh-based garment models, employing networks to predict parameters crucial for clothed human reconstruction. Furthermore, SMPLicit \cite{corona2021smplicit} employs implicit unsigned distance functions to represent the cloth model. Despite yielding promising results, the restricted availability of 3D garment data leads to the unrealism and limited diversity of these cloth models. Recently, some concurrent works \cite{hu2023humanliff,zhang2023text} also attempt to explore multi-layer human generation. HumanLiff\cite{hu2023humanliff} learns layer-wise human generation from 2D images. However, their cloth models are not disentangled with the human body, which limits cloth editing applications.
\cite{zhang2023text} mainly focuses on disentangling the hair and head ornament from the upper body while this paper explores the multi-layer clothes and human body disentangled generation. \cite{wang2023disentangled} proposes generating disentangled clothing using offset mesh construction on the SMPL model. However, due to the topological constraints of SMPL, they are unable to generate loose garment types such as dresses and skirts. In contrast, TELA supports various garment types and can achieve more photorealistic rendering.

\section{Methods}
\begin{figure*}[t]
\centering
\includegraphics[width=1\linewidth,trim={0cm 3cm 0cm 0cm},clip]{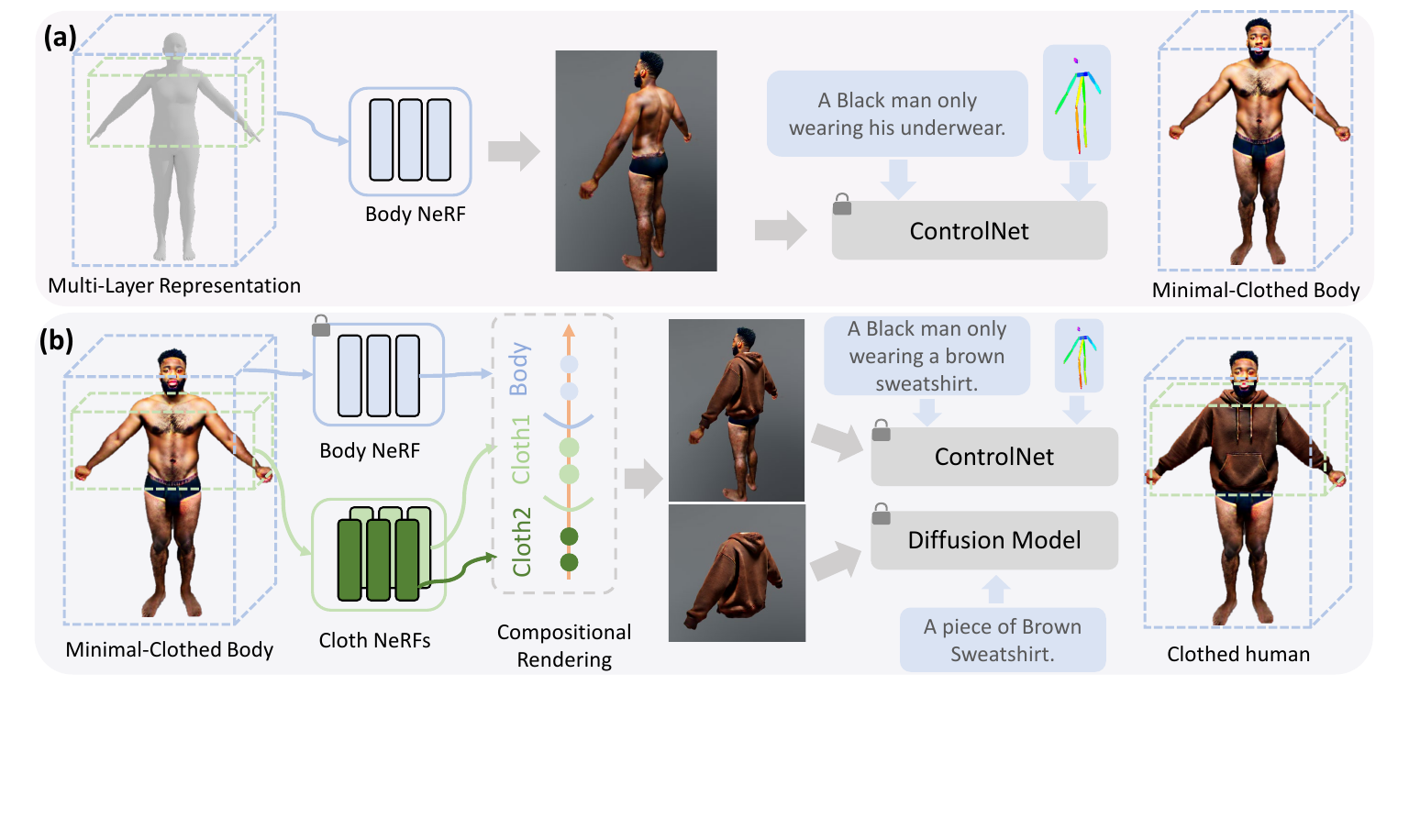}
\caption{\textbf{Overview of TELA.} (a) Minimal-clothed body is the first component to generate. To train the body NeRF, we render the image and the corresponding 2D human skeleton under a random viewpoint and then utilize the 2D human skeleton conditioned ControlNet \cite{controlnet} for SDS optimization. (b) Given the fixed body NeRF, we aim to progressively generate each cloth. For generating cloth $p$, we render an image of the human with cloth $p$ and another image of cloth $p$ only through the proposed transparency-based stratified compositional rendering. Then, the dual SDS losses are proposed to supervise these two images. For the cloth-only image, the original stable diffusion model is adopted for SDS optimization.}
\label{fig:pipeline}
\end{figure*}

Given textual descriptions of a clothed person as input, TELA aims to generate the disentangled 3D human model, where each garment is decoupled from the human body. Figure \ref{fig:pipeline} presents an overview of our approach.
To disentangle each component, we propose a multi-layer clothed human model that is parameterized by multiple neural radiance fields (NeRFs) (Section \ref{sec:multi}) and introduce a novel transparency-based stratified compositional rendering approach (Section \ref{sec:rendering}).
Then, a layer-wise generation method is designed to produce each component based on the text prompts (Section \ref{sec:progressive}).


\subsection{Multi-layer clothed human model}\label{sec:multi}
In contrast to previous methods \cite{huang2023dreamwaltz,kolotouros2023dreamhuman} that encode the holistic clothed human in a single model, we decompose the clothed human into a minimal-clothed human body and multiple clothes (e.g., shirt, coat, and pants). 
The geometry and appearance of clothes are represented as neural radiance fields $F$, which are implemented by MLP networks. 
To better represent human hair, we also employ NeRF as the representation of the minimal-clothed human body rather than the mesh\cite{smpl}.
Each component is represented as a single network. Specifically, for the component $p$, the model can be written as follows:
\begin{equation}
    \sigma(\mathbf{x}), \mathbf{c}(\mathbf{x}) = F_p (h_p(\mathbf{x})),
\end{equation}
where $\sigma(\mathbf{x})$ and $\mathbf{c}(\mathbf{x})$ denote the predicted density and color of the sample point $\mathbf{x}$.
$h_p(\mathbf{x})$ denotes the learnable hash encoding function~\cite{muller2022instant}.

In addition, similar to prior works, we define an `A-pose' human mesh (i.e., SMPL) in the NeRF space to introduce the human prior. Specifically, we utilize the SMPL mesh to determine rough 3D boxes for different clothing, such as upper-body and lower-body clothes. Moreover, during the NeRF optimization, the 3D SMPL skeleton is projected to each viewpoint, and the resulting 2D human skeleton is incorporated with the ControlNet \cite{controlnet} for training (see Section \ref{sec:progressive} for details). 

\subsection{Transparency-based stratified composition} \label{sec:rendering}
Based on the predicted density and color, novel view images can be synthesized through volume rendering:
\begin{gather}
    \tilde{C}(\mathbf{r}) = \sum_{i=1}^{N} T_i (1 - \exp(-\sigma(\mathbf{x}_i) \delta_i)) \mathbf{c}(\mathbf{x}_i), \\
    \text{and} \quad T_i = \exp(-\sum_{j=1}^{i-1} \sigma(\mathbf{x}_j) \delta_j), \label{eq:trans}
\end{gather} 
where $\tilde{C}(\mathbf{r})$ denotes the rendered color of ray $\mathbf{r}$, $\delta_i = || \mathbf{x}_{i + 1} - \mathbf{x}_{i} ||_2$ is the distance between adjacent sampled points, and $T_i$ is the accumulated transparency at sample point $\textbf{x}_i$ along the ray. 
$N$ is the number of sample points along each ray.

Due to the space overlap between different components (e.g., the human body and upper-body clothes), we propose a novel transparency-based stratified compositional rendering method to prevent the penetration between adjacent components. In particular, when training the component $p$, we leverage the inside NeRF networks $\{F_k | k=1,...,p-1\}$ that have been optimized before to predict the densities of sample points along each ray. For each sample point, we take the maximum density value $max\{\sigma_1, ..., \sigma_{p-1}\}$ as its final density. Then, we can calculate the transparency of sample points along each ray with equation \eqref{eq:trans}. \textit{Note that different from previous works \cite{ranade2022ssdnerf,wu2023dorec} using maximum operation to composite multi-layer outputs, we utilize the maximum density for transparency calculation and then divide areas for each layer.} Based on the transparency, we can select sample points whose transparency is larger than a predefined threshold $th$ for training the NeRF network $F_p$. After training, the stratified composition is also utilized to render final clothed human models with multiple NeRFs.
Please also refer to equation (\ref{eq:composition}) for the composition rendering.



\subsection{Training}\label{sec:progressive}

With the multi-layer clothed human representation, we propose a progressive generation framework to obtain each component model sequentially. This framework not only ensures the disentanglement of each component but also provides enhanced flexibility and controllability in the process of clothed human generation. Specifically, we begin by generating a minimal-clothed human body and then progressively generate the clothing outward.




\paragraph{Score Distillation Sampling (SDS).} Before introducing the component generation in detail, we first briefly describe the Score Distillation Sampling \cite{poole2022dreamfusion}. To leverage the supervision from the text descriptions, the SDS loss is proposed to utilize a pre-trained diffusion model $\boldsymbol{\epsilon}_\phi$ for training. Given a 3D model parameterized by $\boldsymbol{\theta}$ and its differentiable rendering image $\mathbf{u} = g(\boldsymbol{\theta})$, the SDS gradients of the 3D model parameters $\boldsymbol{\theta}$ can be written as follows:
\begin{align}
\quad\nabla_{\boldsymbol{\theta}}\mathcal{L}_{\text{SDS}}(\phi,\mathbf{u})=\mathbb{E}_{t, \boldsymbol{\epsilon}}\bigg[w(t)
(\boldsymbol{\epsilon}_\phi(\mathbf{u}_t;y,t)-\boldsymbol{\epsilon})\dfrac{\partial \mathbf{u}}{\partial\boldsymbol{\theta}}\bigg],
\label{eq:ori_sds}
\end{align}
where $w(t)$ denotes a weighting function depending on the timestep $t$, $\mathbf{u}_t$ the noised image, and $y$ the input text prompt. $\boldsymbol{\epsilon}$ is the injected noise added to the rendered image $\mathbf{u}$.

\paragraph{Minimal-clothed body generation.}

The first component to generate is the minimal-clothed body. Similar to previous work \cite{huang2023dreamwaltz}, we adopt the 2D human skeleton conditioned diffusion model to improve the multi-view consistency. Specifically, relying on the predefined `A-pose' SMPL mesh, we can obtain a 3D human skeleton. Then, when the radiance field renders images, the corresponding 2D human skeletons can be obtained by projecting the 3D skeleton to the same viewpoints. Conditioned on the 2D human skeleton $\mathbf{s}$, the SDS loss can be written as follows:
\begin{align}
\quad\nabla_{\boldsymbol{\theta}}\mathcal{L}^{\mathbf{s}}_{\text{SDS}}(\phi,\mathbf{u})=\mathbb{E}_{t, \boldsymbol{\epsilon}}\bigg[w(t)
(\boldsymbol{\epsilon}_\phi(\mathbf{u}_t;y_b,t,\mathbf{s})-\boldsymbol{\epsilon})\dfrac{\partial \mathbf{u}}{\partial\boldsymbol{\theta}}\bigg],
\label{eq:sds}
\end{align}
where $y_b$ denotes the corresponding text prompt of the body.

In addition, we also introduce a regularization loss to prevent floating `radiance clouds': 
\begin{equation} \label{eq:reg}
    L_r= \lambda_{1} \cdot BE(\textbf{M}) + \lambda_{2} \cdot \|\textbf{M}\|_1, 
\end{equation}
where $\textbf{M}$ denotes the rendered mask of the NeRF model. The first item binary entropy function $BE(\cdot)$ encourages the mask to be binarized and the second item introduces the sparsity regularization. $\lambda_{1}$ and $\lambda_{2}$ are constants.

Thus, the full training loss for the minimal-cloth body generation is as follows:
\begin{equation}
    L_b = \mathcal{L}^{\mathbf{s}}_{\text{SDS}} + L_r.
\end{equation}

\paragraph{Cloth generation.}

Given the minimal-clothed body, we aim to progressively generate each garment model. 


To generate the $p$-th piece of clothing, we propose the compositional rendering of the body NeRF $F_b$ and the current cloth NeRF $F_p$. Based on the transparency-based stratified compositional rendering (section \ref{sec:rendering}), we employ the cloth NeRF model $F_p$ to predict the density $\sigma_p$ and color $\mathbf{c}_p$ for those sample points with transparency larger than the threshold $th$ along each ray. For the left sample points, we use the trained body NeRF model $F_b$ to predict corresponding $\sigma_b$ and $\mathbf{c}_b$. This compositional process can be formally written as follows:
\begin{equation} \label{eq:composition}
\begin{split}
\tilde{C}_{bp}(\mathbf{r}) = &\sum_{i=1}^{k} T_i (1 - \exp(-\sigma_p(\mathbf{x}_i) \delta_i)) \mathbf{c}_p(\mathbf{x}_i) + \\ &\sum_{i=k+1}^{N} T_i (1 - \exp(-\sigma_b(\mathbf{x}_i) \delta_i)) \mathbf{c}_b(\mathbf{x}_i),
\end{split}
\end{equation}
where $k$ is the last point with transparency larger than $th$. Intuitively, the above compositional rendering produces an image $\mathbf{u}_{bp}$ of a person wearing the cloth $p$. Based on this image, we can train the cloth model $F_p$ using a loss function similar to $L_b$. Here, we use a text prompt $y_{bp}$ describing a person wearing the cloth $p$ and a new mask $\mathbf{M}_{bp}$.

When only using the compositional rendering images for training, we could ultimately obtain a high-quality model of a person wearing clothes. However, the learned clothing model tends to be coupled with the body model, i.e., the resulting cloth model may include parts of the human body, such as the arms and legs (see Figure \ref{fig:abl_guide}). To address this problem, we introduce an additional cloth-only SDS loss, which encourages the clothing model to only contain the clothing. Specifically, based on the stratified composition, we additionally render an image only using the cloth model as follows:
\begin{gather}
    \tilde{C}_p(\mathbf{r}) = \sum_{i=1}^{k} T_i (1 - \exp(-\sigma_p(\mathbf{x}_i) \delta_i)) \mathbf{c}_p(\mathbf{x}_i).
\end{gather}
Then, we utilize the vanilla SDS loss \eqref{eq:ori_sds} to supervise the cloth-only image $\mathbf{u}_p$ with a text prompt $y_p$ describing the cloth $p$. Furthermore, we also introduce the regularization loss \eqref{eq:reg} for the cloth-only mask $\mathbf{M}_p$. Thus, the full loss for cloth generation is as follows:
\begin{equation} \label{eq:cloth_gene}
    L_p = \mathcal{L}^{\mathbf{s}}_{\text{SDS}}(\mathbf{u}_{bp}; y_{bp}) + L_r(\mathbf{M}_{bp}) + \mathcal{L}_{\text{SDS}}(\mathbf{u}_p; y_p) + L_r(\mathbf{M}_p).
\end{equation}
Note that the parameters of the body model are fixed here.




\section{Experiments}


\paragraph{Implementation details.} We train each NeRF model for $10k$ iterations with batchsize 1, which takes around 3 hours on a single A100 GPU. The NeRF models render images with a resolution of $256 \times 256$ for the first $5k$ iterations and 512x512 for the last $5k$ iterations. Given a text prompt that describes the clothed human, e.g., ``a man wearing cloth $A$, $B$, ...'', we can generate corresponding text prompts for optimization programmatically such as ``a man only wearing underwear'' for mini-clothed body generation, ``a man only wearing cloth A'' and ``a piece of cloth A'' for cloth A generation.


\subsection{Comparisons with the holistic modeling method}

\paragraph{Qualitative comparisons.} To evaluate the quality of the generated clothed human, we compare TELA with the state-of-the-art holistic modeling method DreamWaltz \cite{huang2023dreamwaltz} that introduces 3D consistent score distillation sampling and achieves impressive results. \textit{Please note that while there exist other text-to-human generation approaches, this paper focuses on the cloth-disentangled 3D human generation. These approaches are orthogonal to our work, and our method can also be adapted to their frameworks. } 

\begin{figure*}[htb]
\centering
\includegraphics[width=1\linewidth,trim={0cm 4cm 0cm 3.6cm},clip]{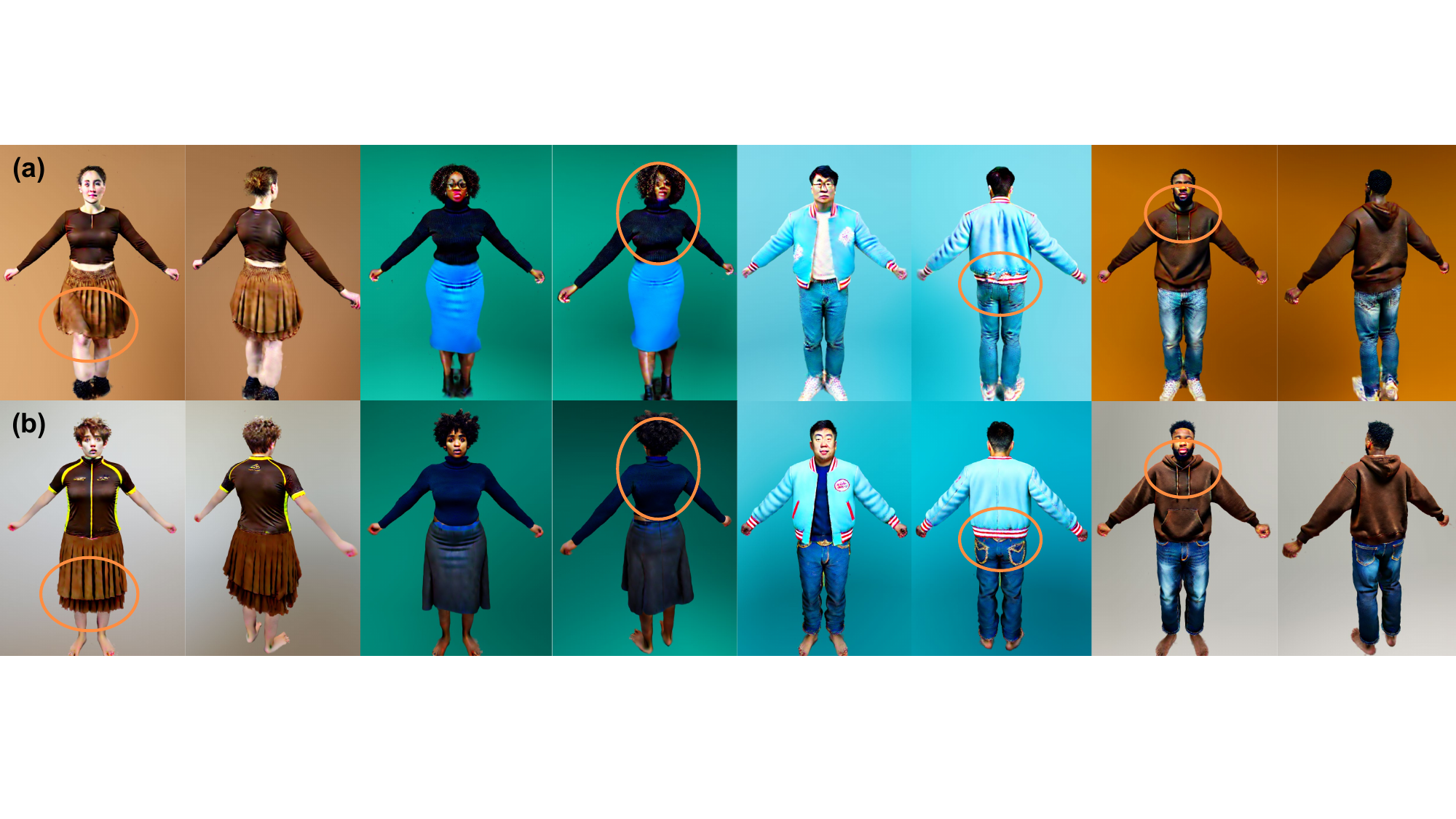} 
\caption{\textbf{Qualitative comparisons with the holistic modeling method \cite{huang2023dreamwaltz}.} (a) DreamWaltz \cite{huang2023dreamwaltz}, (b) Ours. Text prompts (from left to right): ``A woman wearing a Brown Cycling Top and Brown Chiffon Skirt'', ``A Black woman wearing a blue turtleneck and blue Midi Skirt'', ``An Asian man wearing a Light Blue Varsity Jacket and Western Pants'', ``A Black man wearing a Brown Sweatshirt and jeans''}
\label{fig:sota_clothed_human}
\end{figure*}

\begin{figure*}[htb]
\centering
\includegraphics[width=1\linewidth,trim={0cm 4cm 0cm 3.6cm},clip]{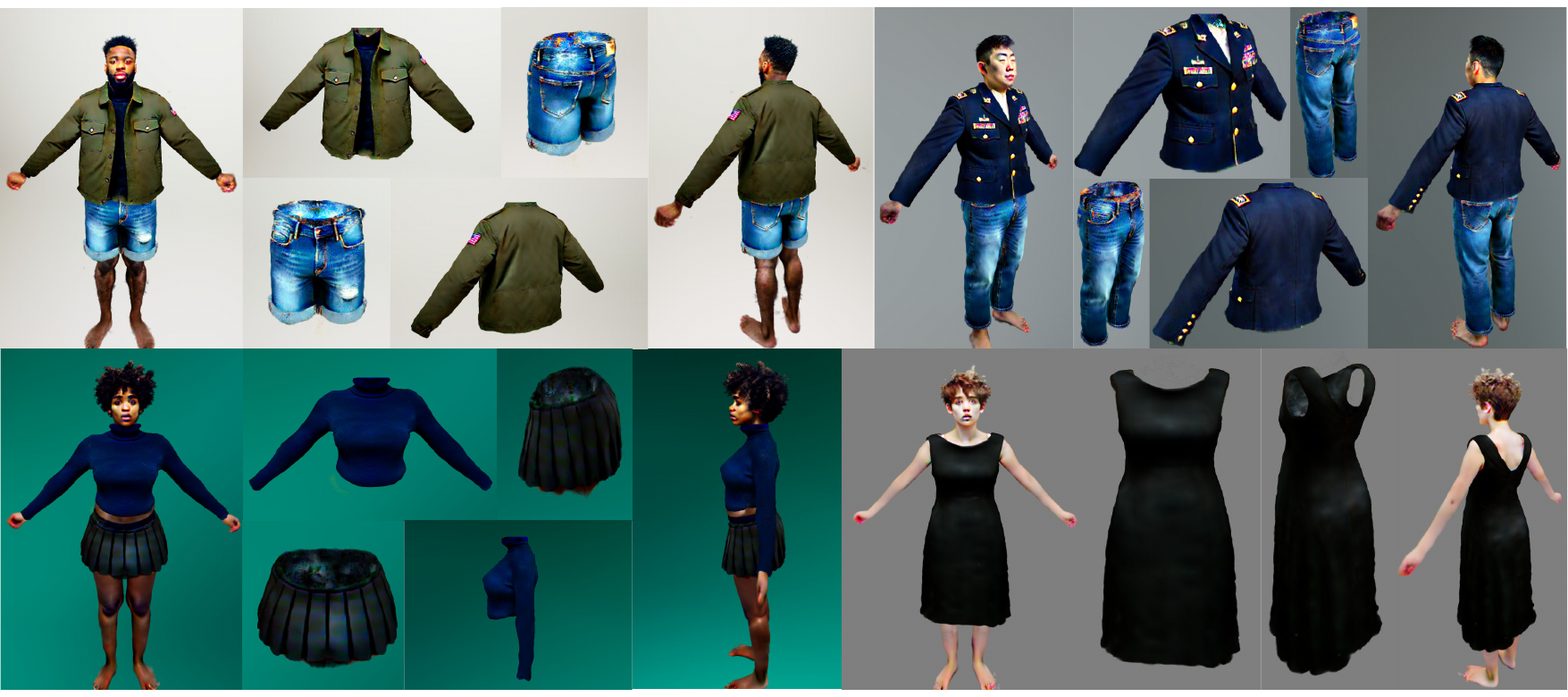}
\caption{\textbf{Qualitative results of the proposed method.} Text prompts: ``A Black man wearing Khaki Outerwear and denim shorts'', ``An Asian man wearing a Navy Blue Military Jacket and jeans'', ``A Black woman wearing a blue turtleneck and Sporty Skirt'', ``A woman wearing Black Dress''.}
\label{fig:qual}
\end{figure*}

We present the qualitative results in Figure \ref{fig:sota_clothed_human}. DreamWaltz \cite{huang2023dreamwaltz} generates clothed humans with blurry and distorted clothes. In contrast, thanks to the disentangled modeling, our method produces high-quality cloth details. Despite we separately model the human body and each cloth, the proposed approach achieves the natural composition of each component. In addition, the results of the second person (i.e., the Black woman) show that disentangled modeling can also help alleviate the multi-face problem. Moreover, we present more qualitative results of clothed humans and disentangled clothes in Figure \ref{fig:qual}.

\begin{figure*}[t]
\centering
\includegraphics[width=1\linewidth,trim={0cm 2.3cm 0cm 3cm},clip]{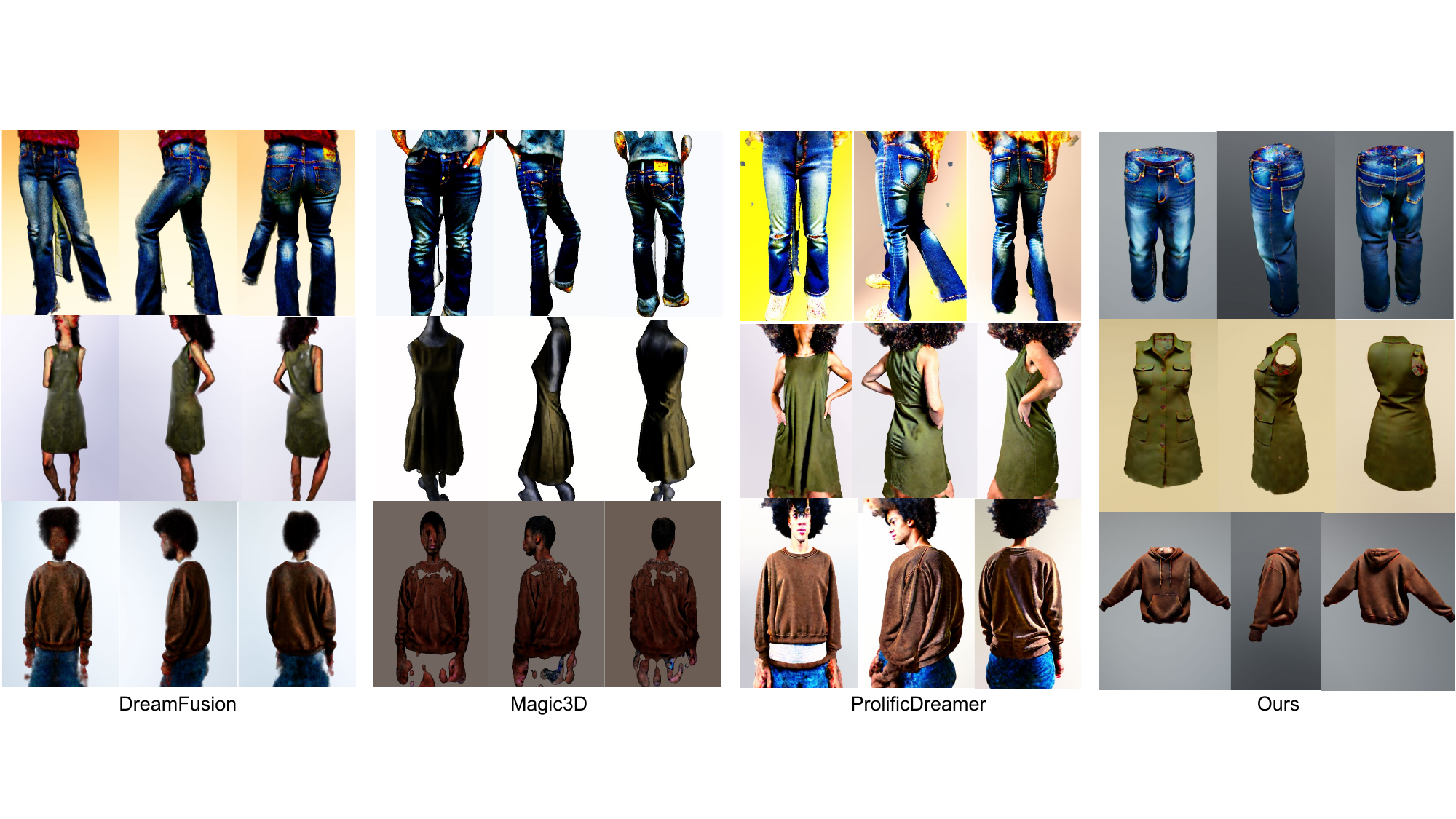}
\caption{\textbf{\textbf{Qualitative comparisons of cloth generation with the SOTA methods.}} Text prompts (from top to bottom): ``a pair of jeans'', ``a piece of Khaki Sleeveless Dress'', ``a Brown Sweatshirt''.}
\label{fig:sota_cloth}
\end{figure*}

\paragraph{Quantitative comparison.}  Quantitative evaluation of generated 3D human models is challenging. We adopt the  Fr\'echet Inception Distance (FID) metric that compares the distribution of two image datasets. We compute the FID between the rendered images of generated models and the images produced by the stable diffusion model. As shown in Table \ref{tb:fid}, we achieve a lower FID score signifying our rendering aligns more closely with the high-quality 2D images from the SD model.

\begin{table}[htb]
\caption{Quantitative comparisons with the holistic modeling method.}\label{tb:fid}

	\begin{center}
		\begin{tabular}{lcc}
                \toprule
			   Methods~~      & DreamWaltz\cite{huang2023dreamwaltz} ~ & Ours  \\
                \midrule
                FID $\downarrow$   & 142.6 & \textbf{125.9}  \\
                \bottomrule
		\end{tabular}
	\end{center}
	\vspace{-0.6cm}
	
\end{table}

\paragraph{User study.} We also conduct user studies to further assess the quality of generated clothed humans against the baseline \cite{huang2023dreamwaltz}. We randomly select 30 text prompts and render generated clothed human models as rotating videos. Given these videos, we asked 26 volunteers to assess the (1) overall clothed human quality, (2) cloth quality, and (3) consistency with text inputs, and selected the preferred results. The quantitative results are shown in Table \ref{tab:user_study}, which demonstrates that the proposed method obtains much higher preference than the baseline over all three metrics.

\begin{table}[htb]
\caption{\textbf{User preference study.} Our method achieves consistently higher preference than the holistic modeling method DreamWaltz \cite{huang2023dreamwaltz} in overall clothed human quality, cloth quality, and consistency with text inputs.}

\label{tab:user_study}    
    \centering
    \begin{tabular}{@{}lc@{}}
   \toprule
    Comparison (Ours vs. DreamWaltz) & Preference (\%) \\
   \midrule
   Overall clothed human quality & 71.46 \\
   Cloth quality & 77.92 \\
   Consistency with text inputs & 73.18 \\
   \bottomrule
\end{tabular}
	\vspace{-0.6cm}

\end{table}

\subsection{Qualitative comparisons of cloth generation} \label{sec:cloth}


Due to the disentangled modeling of the human body and each cloth, our method also enables high-quality 3D cloth generation. To evaluate the quality of generated clothes, we compare it with the state-of-the-art text-to-3D methods: DreamFusion\cite{poole2022dreamfusion}, Magic3D\cite{lin2022magic3d}, and ProlificDreamer\cite{wang2023prolificdreamer}. 

The qualitative results are shown in Figure \ref{fig:sota_cloth}, which presents that the proposed approach significantly outperforms the baselines. First, as shown in the first row (i.e., jeans), even with the view-dependent text augmentation technology \cite{poole2022dreamfusion}, the baselines \cite{poole2022dreamfusion,lin2022magic3d,wang2023prolificdreamer} still struggle for the multi-face problems for the cloth generation. In contrast, by composing with the human body, our method additionally introduces the 2D human skeleton conditioned diffusion model to enhance the multi-view consistency, which dramatically alleviates the multi-face problem. Second, the baselines often generate clothes coupled with the human body while our method enables high-quality disentanglement. Third, thanks to the stratified compositional rendering, as shown in the second row (i.e., sleeve-less dress),  our approach produces clothes with a hollow structure as real clothes, which is crucial for applications like clothing transfer.

\begin{figure}[t]
\centering
\includegraphics[width=0.8\linewidth,trim={0cm 4.5cm 0cm 0cm},clip]{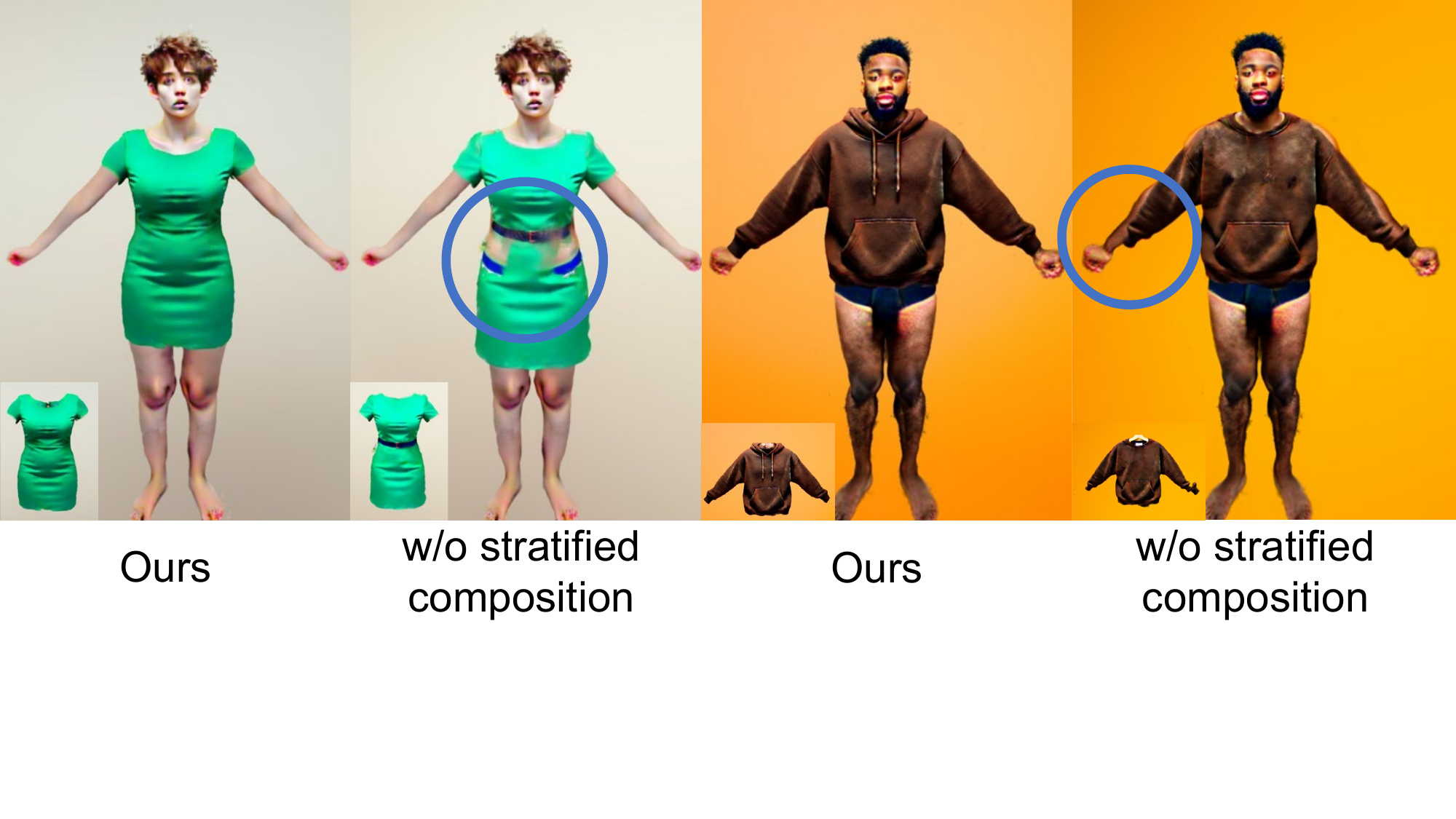}
\caption{\textbf{Ablation studies for Transparency-based stratified compositional rendering.}}
\label{fig:abl_sample}
\end{figure}


\subsection{Ablation studies}
\paragraph{Transparency-based compositional rendering.} As described in Section \ref{sec:multi}, to prevent the penetration between adjacent component NeRFs, Our method utilizes Transparency-based compositional rendering to compose multiple NeRFs. Here, we compare it to the baseline \cite{ranade2022ssdnerf,wu2023dorec} without this compositional rendering, i.e., directly using the maximum operation to fuse the densities of different models and then the corresponding colors are used for volume rendering. The qualitative results of generated clothed human models are presented in Figure \ref{fig:abl_sample}. As shown in the first column, without the proposed compositional rendering, there is serious penetration between the human body and the dress. In addition, as shown in the second column, our compositional rendering can help align clothing models to the human body with the appropriate size.

\begin{figure}[t]
\centering
\includegraphics[width=1\linewidth,trim={0cm 11.2cm 0cm 0cm},clip]{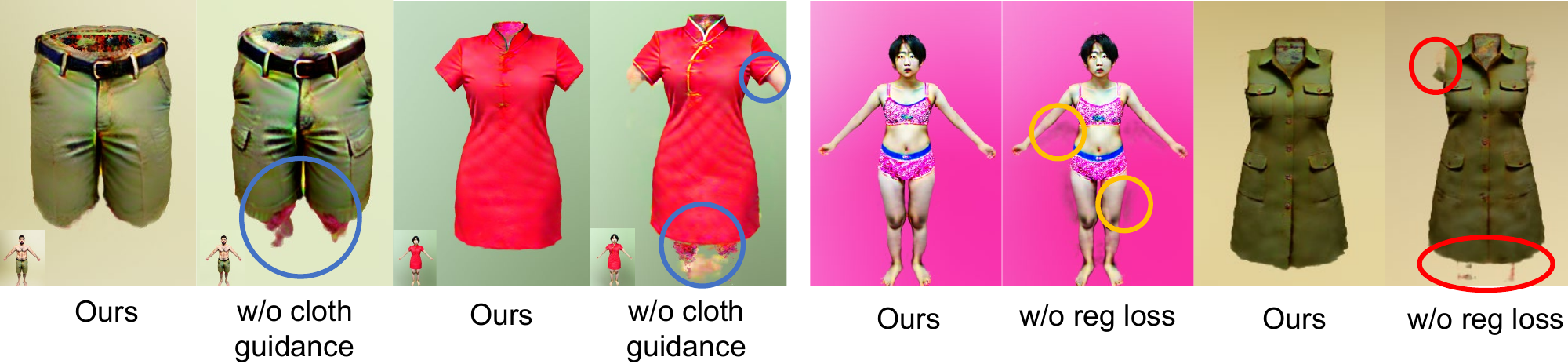}
\caption{\textbf{Ablation studies for cloth guidance loss and regularization loss.}}
\label{fig:abl_guide}
\end{figure}

\paragraph{Dual SDS losses.} As described in Section \ref{sec:progressive}, we propose the dual SDS losses for clothing optimization, i.e., additionally introducing cloth SDS loss. Here, we analyze the effect of the additional cloth SDS loss function. The qualitative results are presented in Figure \ref{fig:abl_guide}. As the results show, when not using the cloth SDS loss, the generated clothing model will entangle with the human body. The reason is that the 2D human skeleton-conditioned SDS loss only supervises the compositional renderings of the clothed human to be consistent with the text input, which does not constrain the cloth model to only contain the clothes. In contrast, the cloth SDS loss provides additional constraints to the cloth model, which can remove the undesired artifacts. Moreover, as shown in section \ref{sec:cloth},  with only cloth SDS loss, the clothing model can not align with the human body and the cloth quality will also degrade. Therefore, the proposed dual SDS losses are effective for high-quality cloth generation.

\paragraph{Regularization loss.} As described in Section \ref{sec:progressive}, we introduce a regularization loss to prevent floating `radiance clouds'. We analyze the impact of this regularization loss, and the results are depicted in Figure \ref{fig:abl_guide}. Removing this loss results in the emergence of floating artifacts around the generated body and clothing.

\subsection{Applications}

\begin{figure}[t]
\centering
\includegraphics[width=0.7\linewidth,trim={2.5cm 0cm 2.5cm 0cm},clip]{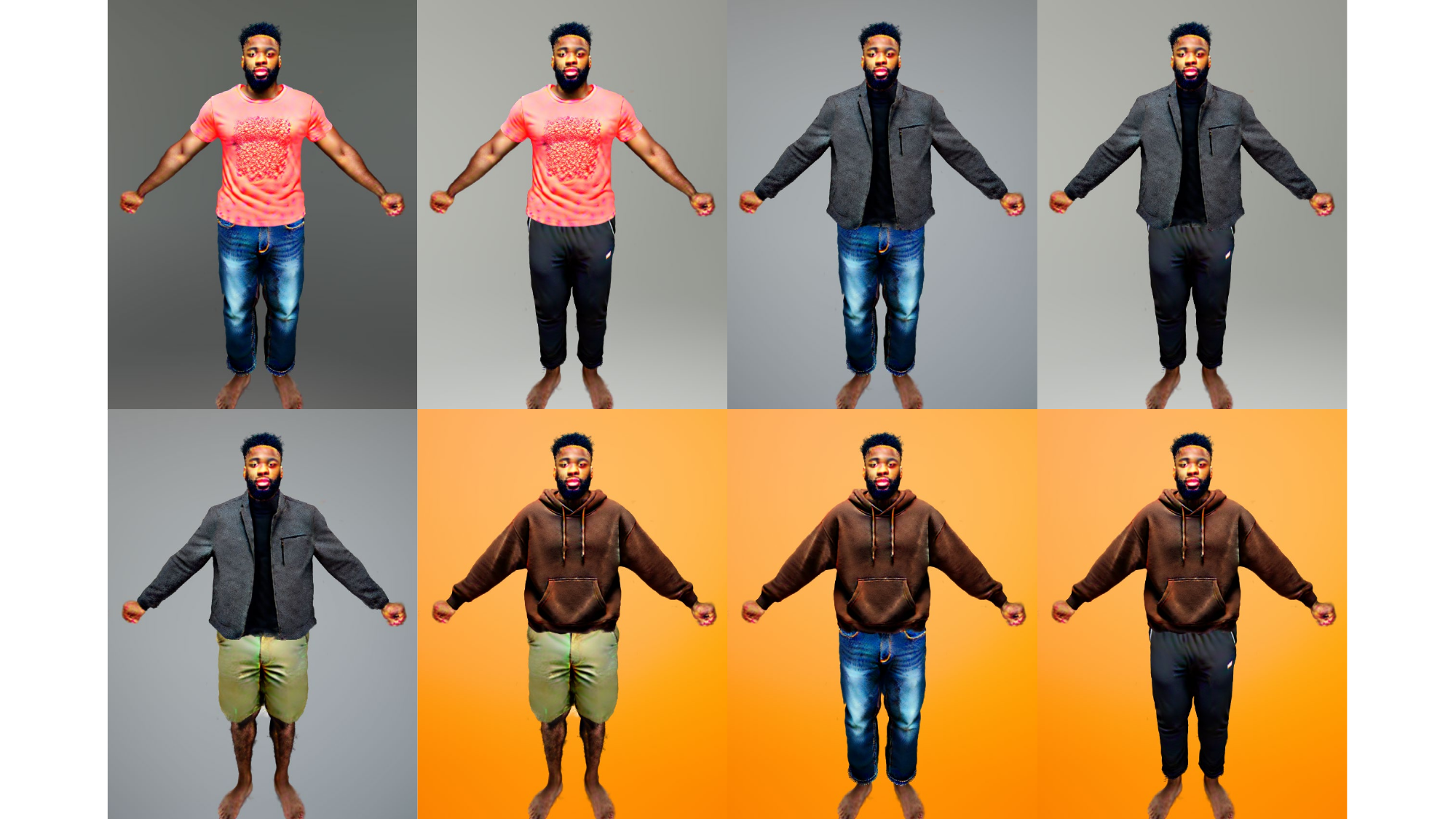}
\caption{\textbf{Free composition of different clothes.}}
\label{fig:free}

\end{figure}

Thanks to the disentangled modeling, our method supports the free composition of different clothes on the same person. We present the free composition of three garments (charcoal gray jacket,  coral T-shirt, and brown sweatshirt) and three pairs of pants (jeans, athletic pants, and khaki shorts) in Figure \ref{fig:free}.

\begin{figure}[t]
\centering
\includegraphics[width=1\linewidth,trim={0cm 4.2cm 0cm 0cm},clip]{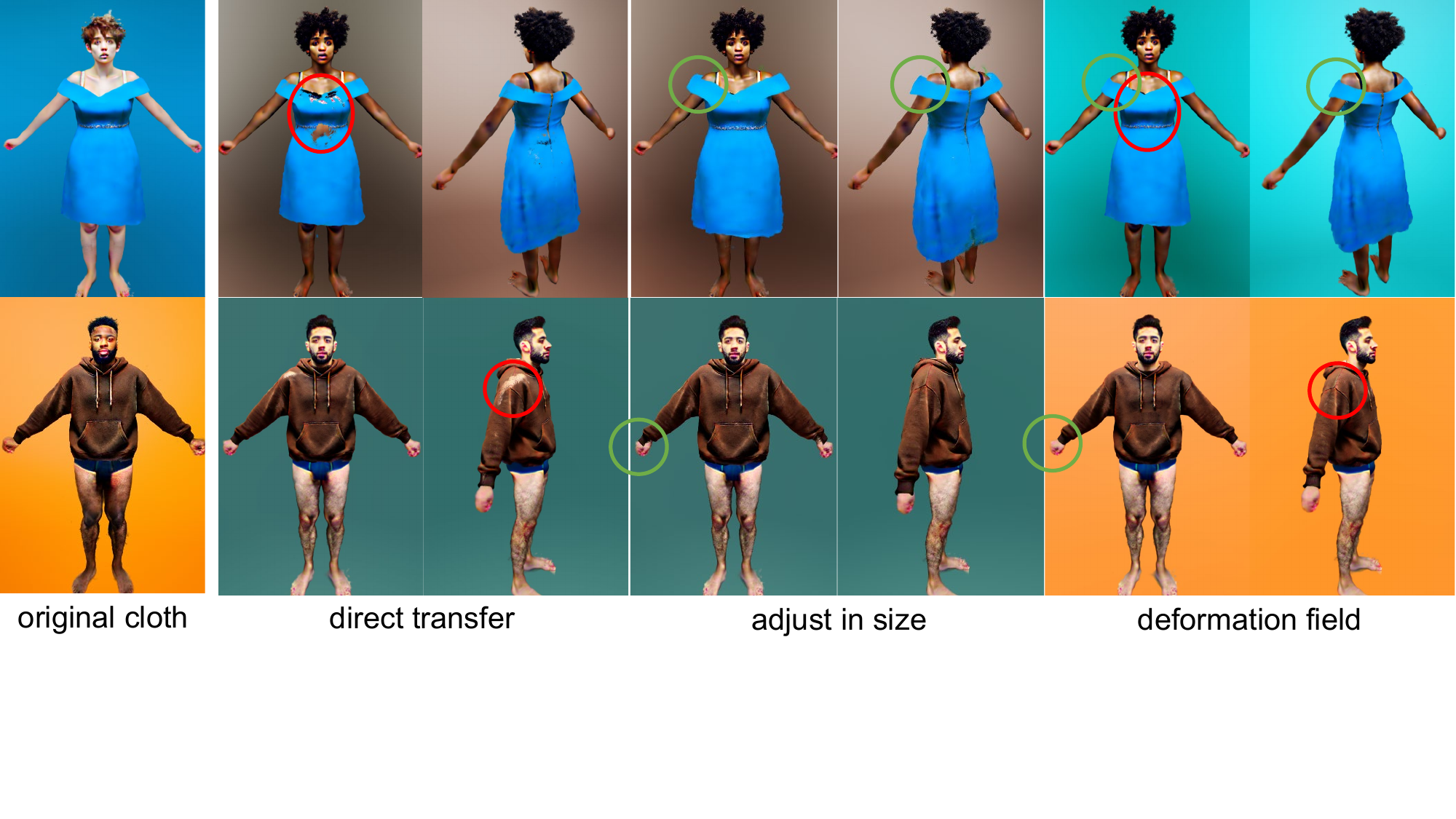} 
\caption{\textbf{Adjusting in size and deformation fields for cloth transfer.} }
\label{fig:trans}
\end{figure}

Additionally, our objective extends to facilitating clothing transfer among individuals with similar body shapes, a valuable pursuit given the clothing sizes similarity. While the NeRF model excels in representing various types of clothing, clothing transfer under NeRF representation poses challenges. \cite{zhang2023text} proposes to adjust the NeRF model in size manually. As shown in Figure \ref{fig:trans}, though they can avoid penetration, the resized clothes do not adapt well to the new person. To solve this, we introduce a clothing deformation field. Specifically, for each sample point $\mathbf{x}$ of the clothing model $p$, we predict its non-rigid deformation using an MLP network $D_p$ and we leverage the same loss functions as cloth generation \eqref{eq:cloth_gene}. The deformation field is trained for $2.5k$ steps. Note that only the parameters of the deformation field are trained here. Based on the deformation field, our method achieves better cloth transfer automatically.

\begin{figure}[t]
\centering
\includegraphics[width=1\linewidth,trim={0cm 8.0cm 0cm 0cm},clip]{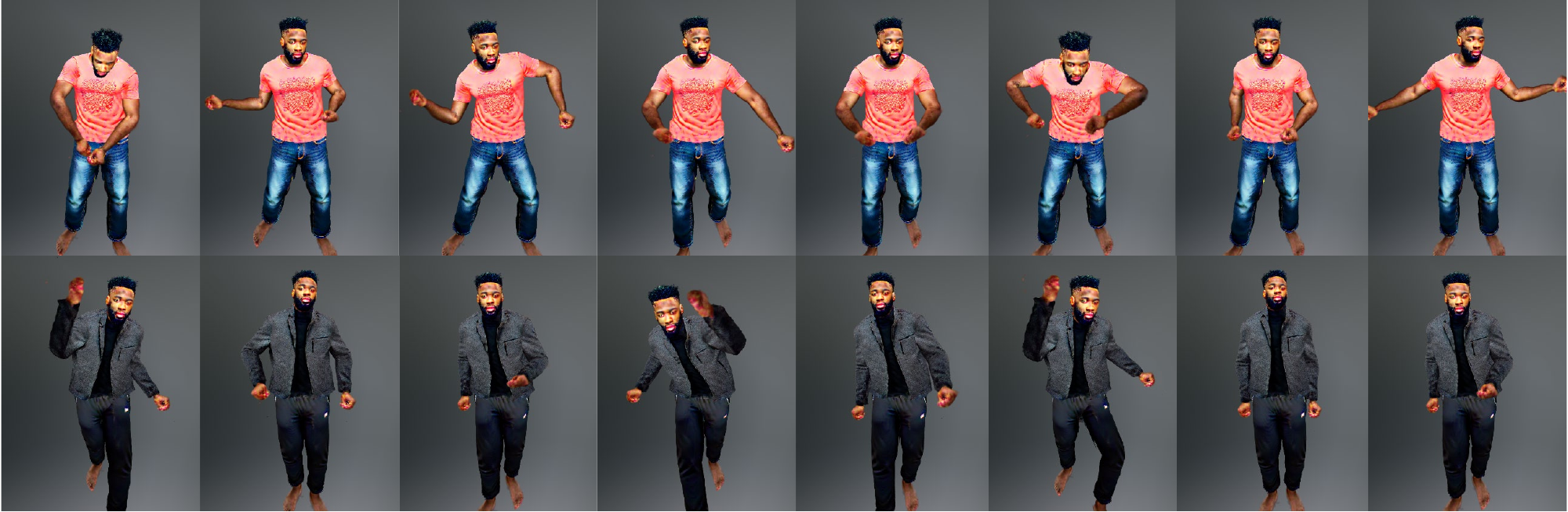} 
\caption{\textbf{Animation results of generated human models.} }
\label{fig:animation}
\end{figure}

Moreover, we can integrate a linear blend skinning module to generate animatable human model. Adopting a similar animation module to DreamWaltz, the proposed method can produce high-quality rendering under novel human poses. The results are presented in Figure \ref{fig:animation}.



\section{Limitations}

The proposed approach still has the following limitations. First, our generation takes hours for optimization, which is much slower than the text-to-image generation. Exploring a more efficient and generalizable model~\cite{liu2023zero1to3} for disentangled human generation could significantly enhance its applicability. Second, while our method could support animatable human generation by introducing the linear blend skinning module like the previous work \cite{huang2023dreamwaltz,kolotouros2023dreamhuman}, it is more interesting to explore clothed human animation with disentangled components (i.e., body and clothes) rather than a holistic model, which is left as future work. Third, the current human body and clothes are represented as NeRFs. The textured mesh representation may be further introduced like \cite{lin2022magic3d}, which can be seamlessly integrated into existing computer graphics systems.


\section{Conclusion}
In summary, this paper presents a novel approach for clothed disentangled 3D human generation from text inputs. Different from previous methods, we propose a multi-layer clothed human representation and progressively generate each component. For superior disentanglement, we first introduce the transparency-based stratified compositional rendering, facilitating the separation of adjacent layers. Then, we propose novel dual SDS losses to help the clothing model decouple from the human body. Thanks to the effective disentanglement, our method enables high-quality 3D garment generation.  Experiments show that our approach not only achieves better clothed human generation but also enables clothing editing applications such as virtual try-on.


\clearpage  

%
%
\bibliographystyle{splncs04}
\bibliography{main}
\end{document}